\documentclass[lettersize,journal]{IEEEtran}
\usepackage{amsmath,amsfonts}
\usepackage{algorithmic}
\usepackage{array}
\usepackage{booktabs}
\usepackage[caption=false,font=footnotesize]{subfig}
\usepackage{hyperref}
\usepackage{cleveref}
\usepackage{float}
\usepackage{multirow}
\usepackage{textcomp}
\usepackage{stfloats}
\usepackage{pdfsync}
\usepackage{url}
\usepackage{verbatim}
\usepackage{graphicx}
\def\BibTeX{{\rm B\kern-.05em{\sc i\kern-.025em b}\kern-.08em
    T\kern-.1667em\lower.7ex\hbox{E}\kern-.125emX}}
\usepackage{balance}

\usepackage{changes}
\definechangesauthor[name=Marco]{MA}
\definechangesauthor[name=Bruno]{BC}

\begin{document}
\title{Experimenting with Normalization Layers in Federated Learning on non-IID scenarios} 
\author{Bruno Casella, Roberto Esposito, Antonio Sciarappa, Carlo Cavazzoni, Marco Aldinucci
\thanks{Bruno Casella, Roberto Esposito and Marco Aldinucci are with the Computer Science Dpt. of the University of Turin, Corso Svizzera 185, Turin, Italy (emails: [bruno.casella, roberto.esposito, marco.aldinucci]@unito.it). \\
Antonio Sciarappa and Carlo Cavazzoni are with Leonardo S.p.A., Italy (emails: antonio.sciarappa.ext@leonardo.com, carlo.cavazzoni@leonardo.com) \\
 \copyright 2022 IEEE. Personal use of this material is permitted. Permission from
IEEE must be obtained for all other uses, in any current or future media,
including reprinting/republishing this material for advertising or promotional
purposes, creating new collective works, for resale or redistribution to servers
or lists, or reuse of any copyrighted component of this work in other works.}}

\markboth{IEEE TRANSACTIONS ON NEURAL NETWORKS AND LEARNING SYSTEMS, VOL. XX, NO. X, XX XXXX}%
{Experimenting with Normalization Layers in Federated Learning on non-IID scenarios}

\maketitle

\begin{abstract}
  Training Deep Learning (DL) models require large, high-quality datasets, often assembled with data from different institutions. Federated Learning (FL) has been emerging as a method for privacy-preserving pooling of datasets employing collaborative training from different institutions by iteratively globally aggregating locally trained models. One critical performance challenge of FL is operating on datasets not independently and identically distributed (non-IID) among the federation participants. Even though this fragility cannot be eliminated, it can be debunked by a suitable optimization of two hyper-parameters: layer normalization methods and collaboration frequency selection. In this work, we benchmark five different normalization layers for training Neural Networks (NNs), two families of non-IID data skew, and two datasets. Results show that Batch Normalization, widely employed for centralized DL, is not the best choice for FL, whereas Group and Layer Normalization consistently outperform Batch Normalization. Similarly, frequent model aggregation decreases convergence speed and mode quality.    
\end{abstract}


\section{Introduction}
\label{sec:introduction}
\IEEEPARstart{T}{he} constant development of Information and Communication Technologies has boosted the availability of computational resources and data, leading us to the Big Data era, where data-driven approaches have become a fundamental aspect of everyday decisions. Both computational resources and data are ubiquitous and inherently distributed. All public and private sectors, from scientific research to companies, take benefit from a vast amount of diverse data to support the growth of their business and to develop more accurate Artificial Intelligence (AI) systems.

Data is often spread and segregated in silos across different institutions and even different business units of the same organization. It is essential to make data accessible to all the partners to train high-quality models and exploit the entire data's value~\cite{kairouz2021advances}. Many recent open science works have encouraged data sharing between institutions in order to improve research possibilities, create collaborations and publish reproducible results. For example, data sharing across countries has been a crucial information tool during the CoViD-19 pandemic~\cite{callaghan2020data}.

However, data is often not shareable due to issues like privacy, security, ownership, trust and economic reasons. For instance, the European regulation GDPR~\cite{voigt2017eu} places stringent constraints on the possibility of sharing sensitive data between parties; industrial companies do not share their data because leveraging it is seen as a competitive advantage. Also, exposing data to other institutions can raise concerns like lack of ownership and lack of trust.

To address these problems, model-sharing strategies (MSS) have emerged as an alternative to data sharing. In MSS, the idea is to share AI models between the involved parties in order to achieve collaboration without sharing raw data. In these approaches, the AI model can range from simpler Machine Learning (ML) algorithms like linear regression to more complex models such as those learned by Deep Learning techniques using Neural Networks (NNs). Recent years have seen the growth of different model-sharing approaches ranging from the "model-to-data remote access" approaches to Federated Learning~\cite{mcmahan2017communication}. In "model-to-data remote access" approaches, AI models are run remotely directly on the machines that hold the data and security is enforced by leveraging secure remote connections and Trusted Execution Enviroments (TEEs) enclaves. Federated Learning has also emerged as a  popular approach. In FL, the involved parties collaborate by aggregating locally trained models into a globally shared one. The process is usually iterative and based on NNs (FedAvg)~\cite{mcmahan2017communication}, even if recently methods based on non-NN distributed boosting algorithms have been proposed ~\cite{polato2022boosting}. These algorithms allow parties to aggregate any kind of model without making assumptions about the kind of model being aggregated or assuming a training procedure based on gradient descent \cite{23:praise-fl:pdp}.

FL is a distributed ML technique originally proposed by Google in 2016 to deal with sensitive data of mobile devices~\cite{mcmahan2017communication}.
FL is an iterative version of model-sharing: clients (the data owners) create a federation (hence the name) together with the server and build a shared model based on the following steps: \emph{1)} clients send their metadata, like the number of classes, training set size, test set size and shape of the input features, to the server, that initializes a model based on the received metadata characteristics; \emph{2)} the server sends the initialized model to all the participants of the federation; \emph{3)} after performing one or more steps of gradient descent, clients send the trained model back to the server; \emph{4)} server acts as an aggregator performing a combination (a function like average, sum, maximum, minimum and so on) of the received models. The aggregated model is now sent to the clients, and steps \emph{3)} and \emph{4)} are repeated until a specified number of rounds are performed, or a convergence criterion is met. The first proposed FL algorithm is the FedAvg~\cite{mcmahan2017communication} algorithm, where the aggregation function used to combine models is the average. In this way, all datasets are kept within the proprietary organizations, and the only information that gets exchanged is the model parameters, that in the case of NN, are matrices of floating point numbers representing the weights and the biases associated with the neurons.

Federated Learning performs well when the data is independently and identically distributed (IID) among the involved institutions. Unfortunately, real-world data is often non-IID, and it is well known that this scenario poses critical issues to FL~\cite{kairouz2021advances}. In a non-IID setting, the data statistics of a single client may be unrepresentative of the global statistics and make the model diverge from the intended solution. Interestingly, Huang at al. show that if the loss surface of the optimization problem is both smooth and convex (which is hardly true in a real scenario), then FedAvg will also converge when the data is non-IID \cite{li2019convergence}.

Recent works have proposed several FL algorithms to cope with non-IIDness problems, such as FedProx~\cite{li2020federated}, FedNova~\cite{wang2020tackling}, SCAFFOLD~\cite{karimireddy2020scaffold}, and FedCurv~\cite{shoham2019overcoming}, which has been tested in \cite{li2022federated, casella2022benchmarking}. Notice that all these algorithms are modified versions of FedAvg and they preserve the principle underneat FedAvg: to average the weights in all the layers of the NN. Most of the common NN architectures employ Batch Normalization (BN)~\cite{ioffe2015batch}, a technique for improving the training of NNs to make them converge faster and to a more stable solution. BN works by standardizing the layers' input for each mini-batch. 

In this work, we investigate two aspects of the training FL models which, differently from cenytralized case, they happen to be hyper-parameters that can be optimized: the normalization layers and the frequency of models aggregation (epochs per round). We show that the most popular normalization layer (BN) does not couple well with FL for non-IID data and that substituting BN with alternative normalization FL a better model can be produced for both the non-IID and IID case. We also show that building a global model aggregating local models at each epoch is not a good strategy, neither for quality of the model nor for the execution time. We  experiment with two network architectures and five different normalization layers on two public image datasets: MNIST~\cite{lecun1998gradient} and CIFAR-10~\cite{krizhevsky2009learning}.
Results show that the performance of the networks is strictly related to the type of normalization layer adopted.  

The main contributions of this work are:
\begin{itemize}
  \item We provide benchmarks for five different normalization layers: BN, GN, LN, IN, BRN;
	\item We provide results of experiments on FedAvg on two non-IID settings considering a feature distribution skew and a labels distribution skew (in addition to the IID case). To the best of our knowledge, this is the first work providing empirical evidence on the behaviour of these normalization layers in common non-IID cases;
	\item for the most promising normalization layers, we ran extensive tests to discuss how performances are affected by the following factors:
	\begin{enumerate}
		\item Batch size.
		\item Number of epochs per round (E).
		\item Number of clients.
	\end{enumerate}
 	\item We show that choosing the right normalization layer and a suitable number of local gradient descent steps is crucial for obtaining good performances. 
\end{itemize}
This work extends the typical search for optimization of machine learning parameters to federated learning.

The rest of the paper is organized as follows. In Section~\ref{sec:related-work}, we introduce and discuss recent related works. In Section~\ref{sec:normalization-layers}, the most used normalization layers are reviewed. In Section~\ref{sec:non-IID-data}, the most typical non-IID scenarios are described. Section~\ref{sec:experiments} shows and discusses experimental results. Finally, Section~\ref{sec:conclusions} concludes the paper.

\section{Related Work}
\label{sec:related-work}
\noindent The main challenges in FL are statistical heterogeneity (non-iidness) and systems heterogeneity (variability of the devices of the federation). In this work, we address the former. In \cite{kairouz2021advances}, the most common non-IID data settings that are quantity skew, labels quantity skew (prior shift), feature distribution skew (covariate shift), same label but different features and same features but different labels, are reviewed. To the best of our knowledge, there are only a few benchmarks for FL dealing with non-IID data. Li et al. in \cite{li2022federated} report the analysis of FedAvg~\cite{mcmahan2017communication}, FedNova~\cite{wang2020tackling}, FedProx~\cite{li2020federated} and SCAFFOLD~\cite{karimireddy2020scaffold} on nine public image datasets, including MNIST~\cite{lecun1998gradient} and CIFAR10~\cite{krizhevsky2009learning}, split according to three of the previous mentioned non-IID partition strategies, i.e. quantity skew, labels quantity skew and three different versions of feature distribution skew: noise-based, synthetic and real-world feature imbalance. Authors show that none of those algorithms outperforms others in all the cases and that non-iidness degrades the performance of FL systems in terms of accuracy, especially in the case of labels quantity skew. Another recent work~\cite{casella2022benchmarking} reports an empirical assessment of the behaviour of FedAvg and FedCurv~\cite{shoham2019overcoming} on MNIST, CIFAR10 and MedMNIST~\cite{yang2021medmnist}. Datasets are split according to the same non-IID settings of \cite{li2022federated}. Authors show that aggregating models at each epoch is not necessarily a good strategy: performing local training for multiple epochs before the aggregation phase can significantly improve performance while also reducing communication costs. FedAvg produced better models in most non-IID settings despite competing with an algorithm explicitly developed to deal with these scenarios (FedCurv). 

Results in \cite{casella2022benchmarking} also confirmed literature sentiment: labels quantity skew and its pathological variant are the most detrimental ones for the algorithms. The same non-IID partitions have already been tested in \cite{polato2022boosting}, which proposes a novel technique of non-gradient-descent FL on tabular datasets.
Our paper extends \cite{casella2022benchmarking}, deepening the experiments about the number of epochs per round, a hyper-parameter that, if tuned appropriately, can lead to large performance gains. Moreover, we aim to investigate which type of normalization layer better fits FL on non-IID data. Indeed, when data are non-IID, batch statistics do not represent the global statistics, leading NNs equipped with BN to poor results. The most common alternatives to BN are: Group Normalization (GN)~\cite{wu2018group}, Layer Normalization (LN)~\cite{ba2016layer}, Instance Normalization (IN)~\cite{ulyanov2016instance} and Batch Renormalization (BRN)~\cite{ioffe2017batch}. To the best of our knowledge, there are no works benchmarking normalization layers for FL on non-IID data. A previous work \cite{casella2022transfer}, proposing a novel form of Transfer Learning through test-time parameters' aggregation, shows that a NN with Batch Normalization~\cite{ioffe2015batch} does not learn at all, while performance improves only when using Group Normalization~\cite{wu2018group}. Andreaux et al. propose a novel FL approach by introducing local-statistic BN layers \cite{andreux2020siloed}. Their method, called SiloBN, consists in only sharing the learned BN parameters $\gamma$ and $\beta$ across clients, while BN statistics $\mu$ and $\sigma^2$ remain local, allowing the training of a model robust to the heterogeneity of the different centres. SiloBN showed better intra-centre generalization capabilities than existing FL methods. FedBN~\cite{li2021fedbn} is a FL algorithm that excludes BN layers from the averaging step, outperforming both FedAvg and FedProx in a feature distribution skew setting.

\section{Normalization Layers}
\label{sec:normalization-layers}
\noindent The majority of the FL algorithms simply apply an aggregation function (like averaging) to all the components of a NN, including weights and biases of the normalization layers. Most of the common NN architectures, like residual networks~\cite{he2016deep}, adopt BN~\cite{ioffe2015batch} as normalization layer. However, in contexts like Federated or Transfer Learning, BN may not be the optimal choice, especially when dealing with non-IID data. In this chapter will be reviewed the main characteristics of Batch Normalization and several possible alternatives like Group Normalization (GN)~\cite{wu2018group}, Layer Normalization (LN)~\cite{ba2016layer}, Instance Normalization (IN)~\cite{ulyanov2016instance} and Batch Renormalization (BRN)~\cite{ioffe2017batch}.
\subsection{Batch Normalization}
\label{ssec:batch-norm}
Batch Normalization has seen a recent extensive adoption by neural networks for their training. The key issue that BN tackles is Internal Covariate Shift (ICS), which is the change in the distribution of the data (or network activations), i.e. the input variables of training and test sets. Informally, at each epoch of training, weights are updated, input data are different, and the algorithm faces difficulties. This results in a slower and more difficult training process because lower learning rates and careful parameter initialization are then required. BN attempts to reduce ICS by normalizing activations to stabilize the mean and variance of the layer's inputs. This accelerates training by allowing the use of higher learning rates and reduces the impact of the initialization. During training, BN normalizes the output of the previous layers along the batch size, height and width axes to have zero mean and unit variance:
$$\hat{x}_i = \frac{x_i - \mu_m}{\sqrt{\sigma^2_m + \epsilon}}$$
where $x, \mu_m$ and $\sigma^2_m$ are respectively the input, the mean and the variance of a minibatch $m$, and $\epsilon$ is arbitrarily constant greater than zero used for stability in case the denominator is zero. BN also adds two learnable parameters, $\gamma$ and $\beta$ that are a scaling and a shifting step, to fix the representation in case the normalization alters what the layer represents: $y_i = \gamma \hat{x_i} + \beta$. Normalized activations will depend on the other samples contained in the minibatch. In the test phase, BN can not calculate statistics; otherwise, it will learn from the test dataset, so it uses the moving averages of minibatch means and variances of the training set. In the case of IID mini-batches, statistical estimations will be accurate if the batch size is large; otherwise, inaccuracies will be compounded with depth, reducing the quality of the models. Non-IID data can have a more detrimental effect on models equipped with BN because batch statistics do not represent global statistics, leading to even worse results. Therefore there is a need to investigate alternatives to BN that can work well with non-IID data and small batch sizes.
\subsection{Group Normalization}
\label{ssec:group-norm}
Group Normalization is a simple alternative to BN. It divides the channels into different groups and computes within each group the mean $\mu_i$ and the variance $\sigma_i$ along the height and width axes. GN overcomes the constraint on the batch size because it is completely independent of the other input features in the batch, and its accuracy is stable in a wide range of batch size. Indeed, GN has a 10.6$\%$ lower error than BN on ResNet-50~\cite{he2016deep} trained on ImageNet~\cite{wu2018group}. 
The number of groups G is a pre-defined hyperparameter which needs to divide the number of channels C. When G=C, it means that each group contains one channel, and GN becomes Instance Normalization, while when G=1, it means that one group contains all the channels, and GN becomes Layer Normalization. Both Instance and Layer Normalizations are described below.
\subsection{Instance Normalization}
\label{ssec:instance-norm}
Instance Normalization is another alternative to BN, firstly proposed for improving NN performances in image generation. It can be seen as a Group Normalization with G=C or as a BN with a batch size of one, so applying the BN formula to each input feature individually. Indeed, IN computes the mean $\mu_i$ and the variance $\sigma_i$ along the height and width axes. As stated before, BN suffers from small batch sizes, so we expect that experiments made with IN will produce worse results than the ones with BN or GN, which can exploit the dependence across the channels.
\subsection{Layer Normalization}
\label{ssec:layer-norm}
Layer Normalization was first proposed to stabilize hidden state dynamics on Recurrent Neural Networks (RNNs)~\cite{ba2016layer}. It computes the mean and the variance along the channel, height and width axes. LN overcomes the constraint on the batch size because it is completely independent of the other input features in the batch. LN performs the same computation both at training and inference times. It can be seen as a GN with G=1, so with only one group controlling all the channels. As a result, when there are several distributions to be learned among the group of channels, it can perform worse than GN.
\subsection{Batch Renormalization}
\label{ssec:batch-renorm}
Batch Renormalization~\cite{ioffe2017batch} is an extension of BN that ensures training and inference models generate the same outputs that depend on individual examples rather than the entire minibatch. BRN is an augmentation of a network which contains batch normalization layers with a per-dimension affine transformation applied to the normalized activations to ensure the match between training and inference models. Reducing the dependence of activation of each sample with other samples in the minibatch can result in a performance increase when data are non-IID.

\section{Non-IID Data}
\label{sec:non-IID-data}
\noindent The most common non-IID data settings are reviewed in \cite{kairouz2021advances} that lists five different partitioning strategies: 1) quantity skew, 2) labels quantity skew (prior shift), 3) feature distribution skew (covariate shift), 4) same labels but different features and 5) same features but different labels. In this paper, we consider the same distributions tested in \cite{casella2022benchmarking, li2022federated} apart from quantity skew, which is not treated. Indeed, \cite{casella2022benchmarking, li2022federated} showed that quantity skew does not hurt the performance of FL models, probably because it results in a different quantity of samples per client, but the distribution of samples is uniform, which is easy to deal with. In this paper, labels quantity skew, which is the most detrimental to the FL models' performance, has been extensively tested in a lot of scenarios to show how it is possible to overcome its difficulties. The cases adopted (both IID and non-IID) are briefly described.
\begin{itemize}
	\item \textbf{Uniform Distribution (IID)}: each client of the federation holds the same amount of data, and the distribution is uniform among parties. This is the simplest case for FL algorithms because the distribution is IID.
	\item \textbf{Labels Quantity Skew}: the marginal distributions of labels $P(y_i)$ vary across parties, even if $P(x_i|y_i)$ is the same. This especially happens when dealing with real FL applications where clients of the federation are distributed among different world regions. Certain data are present only in some countries, leading to the labels quantity skew. In this work, we adopted the simplest version of labels quantity skew, where each client holds samples belonging to only a fixed amount of classes. In our experiments, we used two as number of classes per client. Other versions of labels quantity skew can be the Dirichlet labels skew (each client holds samples such that classes are distributed according to the Dirichlet function) and the Pathological labels skew (data are firstly sorted by label and then divided in shards). 
	\item \textbf{Feature Distribution Skew}: the marginal distributions $P(x_i)$ vary across parties, even if $P(y|x)$ is shared. This can happen in a lot of ML scenarios; for example, in handwriting recognition, the same words can be written with different styles, stroke widths, and slants. The covariate shift was obtained according to the procedure described in \cite{polato2022boosting}: samples are distributed among clients according to the results of a Principal Component Analysis (PCA) performed on the data.
\end{itemize}

\section{Experiments}
\label{sec:experiments}
\noindent Our experiments have been realized using OpenFL~\cite{reina2021openfl}, the new framework for FL developed by Intel Internet of Things Group and Intel Labs. OpenFL is a Python 3 library for FL that enables organizations to collaboratively train a model without sharing sensitive information. OpenFL is DL framework-agnostic. Training of statistical models may be done with any deep learning framework, such as TensorFlow or PyTorch, via a plugin mechanism.
OpenFL is based on a Director-Envoy workflow which uses long-lived components in a federation to distribute more experiments in the federation. The Director is the central node of the federation. It starts an Aggregator for each experiment, sends data to connected collaborator nodes, and provides updates on the status. The Envoy runs on Collaborator nodes connected to the Director. When the Director starts an experiment, the Envoy starts the Collaborator to train the global model.
All the experiments were computed in a distributed environment with ten collaborators. Each collaborator is run on an Intel Xeon CPU (8 cores per CPU), 1 Tesla T4 GPU. The code used for experimental evaluation is publicly available at \href{https://github.com/alpha-unito/Benchmarking-Normalization-Layers-in-Federated-Learning-for-Image-Classification-Tasks-on-non-IID}{https://github.com/alpha-unito/Benchmarking-Normalization-Layers-in-Federated-Learning-for-Image-Classification-Tasks-on-non-IID}.

\noindent\textbf{Dataset}: We tested FedAvg on MNIST~\cite{lecun1998gradient} and CIFAR10~\cite{{krizhevsky2009learning}}, that are default benchmarks in ML literature. The details of the datasets are summarized in Table \ref{tab:dataset-stats}.

\begin{table}
\centering
\caption{\label{tab:dataset-stats}Statistics of the datasets.\\}
\label{tab:freq}
\begin{tabular}{llll}
\toprule
\textbf{Dataset}  & \textbf{Train samples} & \textbf{Test samples} &  \textbf{\# labels} \\
\midrule
MNIST    & 60.000        & 10.000       & 10        \\
\midrule
CIFAR10  & 50.000        & 10.000       & 10        \\
\bottomrule
\end{tabular}
\end{table}

\noindent\textbf{Preprocessing}: both datasets were not rescaled: MNIST images are 28x28 while CIFAR10 images are 32x32. As for data augmentation, we performed random horizontal flips and random crop with a probability of 50\%. All the datasets were normalized according to their mean and standard deviation.

\noindent\textbf{Model}: We employed ResNet-18~\cite{he2016deep} and EfficientNet-B0~\cite{tan2019efficientnet} as classification models, trained by minimizing the cross-entropy loss with mini-batch gradient descent using the Adam optimizer with learning rate $10^{-3}$. The local batch size was 128. We used two networks to show that the results are not model-dependent (See \ref{sec:appendix} for the EfficientNet-B0's results). The scores of baseline models and federated experiments on the uniform and non-IID settings (section. \ref{ssec:norm-layers-non-IID}, \Cref{tab:non-federated,tab:LABELSSKEW-setting,tab:COVARIATE-setting}) are the average ($\pm$ standard deviation) over five runs. For the extensive experiments on batch size, number of local training steps, and number of clients, we tested only ResNet-18 for only one run.

\noindent\textbf{Normalization Layers}: All the normalization layers described before, i.e. Batch Norm, Group Norm, Instance Norm, Layer Norm and Batch Renormalization, have been applied to the classification model in each experiment. For the most promising normalization layers have been ran additional tests to study the impact of the batch size, the number of epochs per round and the number of clients. For BN, we set the momentum, i.e. the importance given to the previous moving average, to 0.9, according to the SOTA~\cite{moreau2022benchopt} for ResNet-18. For GN, the number of channels must be divisible by the number of groups, so we set the number of groups to 32 for ResNet-18 (one of the possible divisors) and 8 for EfficientNet-B0 (the only possible divisor). All the other normalization layers have been used with their standard PyTorch configuration.

\begin{figure*}[!ht]
  \subfloat[BN\label{subfig-1:epochs1}]{%
    \includegraphics[width=0.33\textwidth]{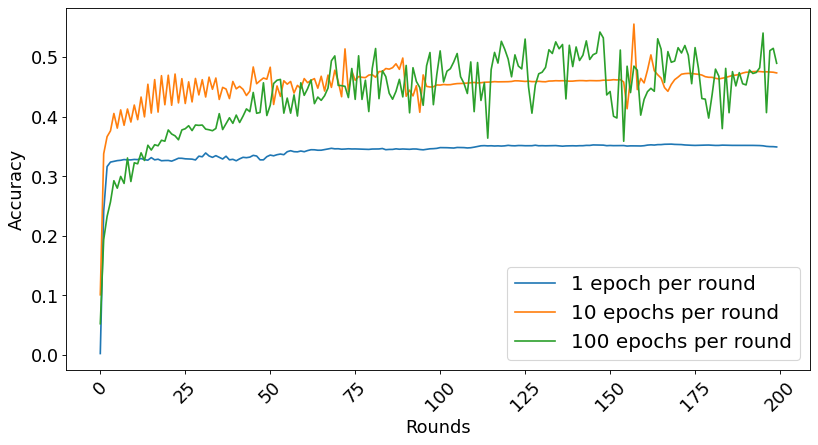}
  }
  \hfill
  \subfloat[GN\label{subfig-2:epochs10}]{%
    \includegraphics[width=0.33\textwidth]{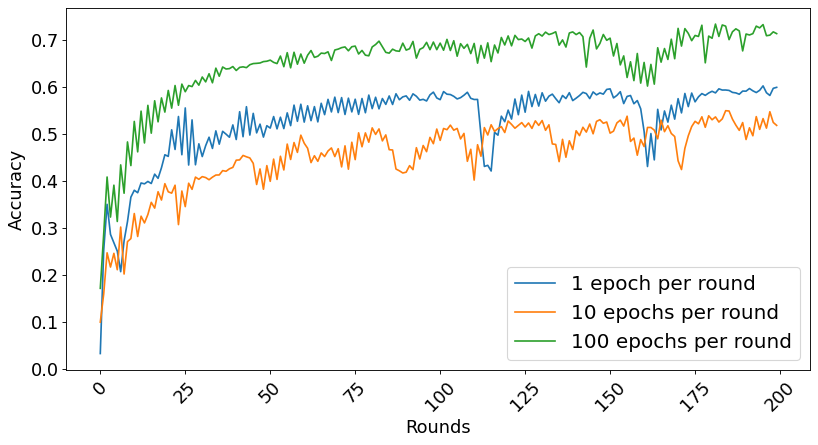}
  }
  \subfloat[LN\label{subfig-3:epochs100}]{%
    \includegraphics[width=0.33\textwidth]{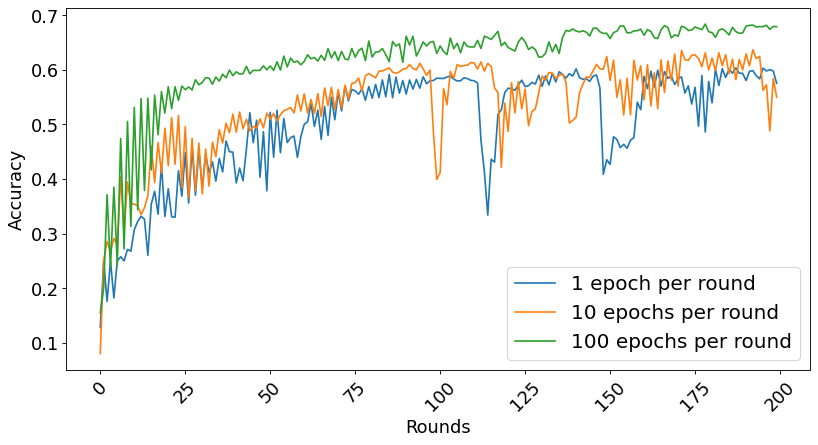}
  }
  \caption{\textbf{Accuracies on CIFAR10 and different epochs per round.} Accuracy increases as BN converges lates as E increases, while GN and LN follow an inverse pattern.}
  \label{fig:epochs}
\end{figure*}

Top-1 accuracy has been employed as a classification metric to compare the performance. Results show the best aggregated model's accuracy. The learning curve of all the experiments can be studied from \Cref{fig:non-iid} to \Cref{fig:CLIENTS}. Table \ref{tab:non-fed-setting} reports about a non-federated baseline, i.e., the typical AI scenario where the data are centralized. The remaining tables show the performance of FedAvg in different data partitioning scenarios and for different values of some hyperparameters such as batch size, number of epochs per round and number of clients.

\begin{table*}
\centering
\caption{\label{tab:non-fed-setting}Accuracy in the non-federated setting.\\}
\label{tab:non-federated}
\begin{tabular}{lccccccc} 
\toprule
\textbf{Dataset}            & \textbf{BN}  & \textbf{GN}  & \textbf{IN} & \textbf{LN} & \textbf{BRN}\\ 

\midrule
MNIST        & $99.22\% \pm 0.07\%$          &   $99.29\% \pm 0.06\%$      &    $11.36\% \pm 0.00\%$       &    $\textbf{99.32\%} \pm \textbf{0.07\%}$        &    $99.00\% \pm 0.20\%$       \\
\midrule
CIFAR10      & $\textbf{82.60\%} \pm \textbf{0.24\%}$          &    $81.61\% \pm 0.24\%$     &    $9.89\% \pm 0.00\%$        &    $82.06\% \pm 0.40\%$        &    $81.77\% \pm 0.23\%$                 \\
\bottomrule
\end{tabular}
\end{table*}

\subsection{Normalization Layers and non-IID data}
\label{ssec:norm-layers-non-IID}
\begin{figure*}[!ht]
  \subfloat[\normalfont\footnotesize{IID MNIST}\label{subfig-1:uniform}]{%
    \includegraphics[width=0.33\textwidth]{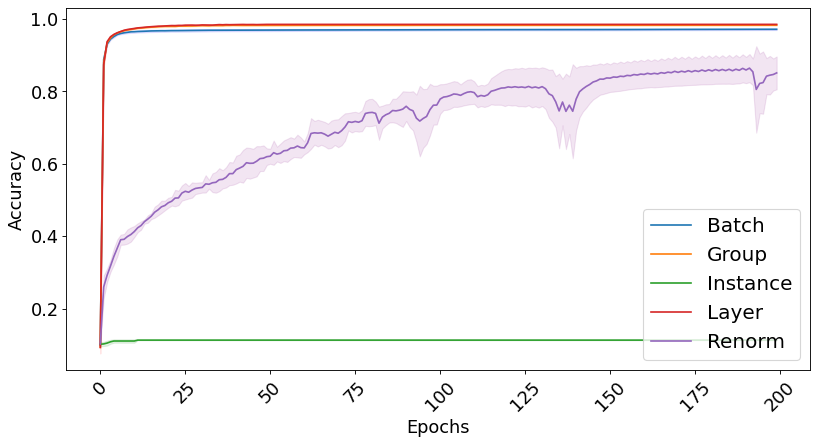}
  }
  \subfloat[\normalfont\footnotesize{Non-IID (labels quantity skew) MNIST}\label{subfig-2:labels}]{%
    \includegraphics[width=0.33\textwidth]{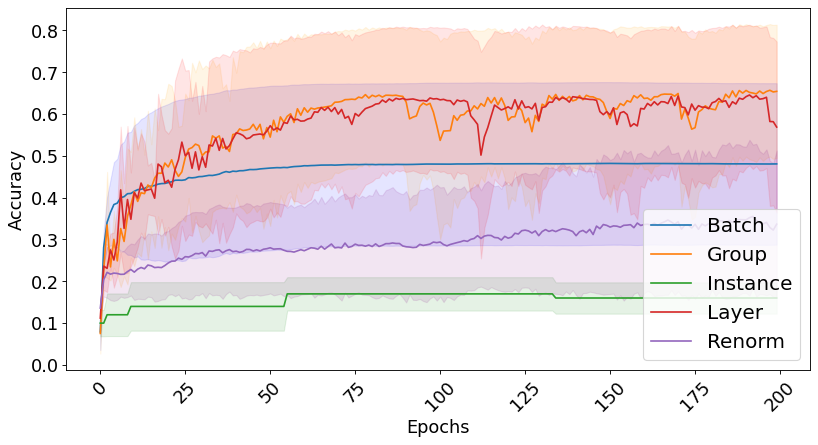}
  }
  \subfloat[\normalfont\footnotesize{Non-IID (covariate shift) MNIST}\label{subfig-3:covariate}]{%
    \includegraphics[width=0.33\textwidth]{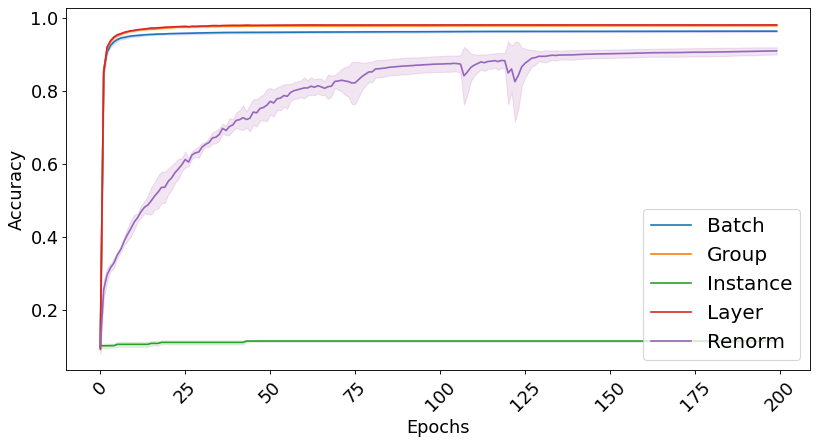}
  }
  \hfill
  \subfloat[\normalfont\footnotesize{IID CIFAR10}\label{subfig-1:cifaruniform}]{%
    \includegraphics[width=0.33\textwidth]{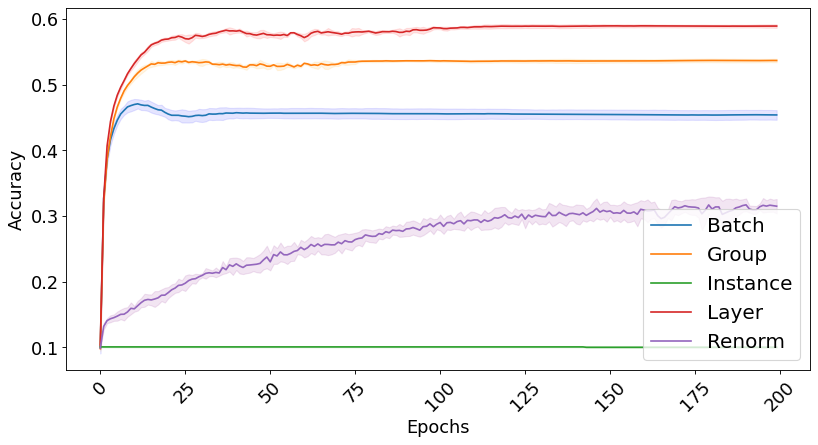}
  }
  \subfloat[\normalfont\footnotesize{Non-IID (labels quantity skew) CIFAR10}\label{subfig-2:cifarlabels}]{%
    \includegraphics[width=0.33\textwidth]{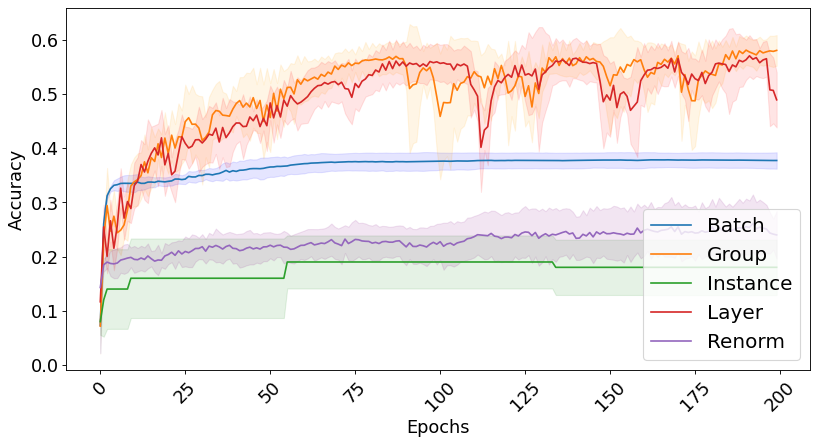}
  }
  \subfloat[\normalfont\footnotesize{Non-IID (covariate shift) CIFAR10}\label{subfig-3:cifarcovariate}]{%
    \includegraphics[width=0.33\textwidth]{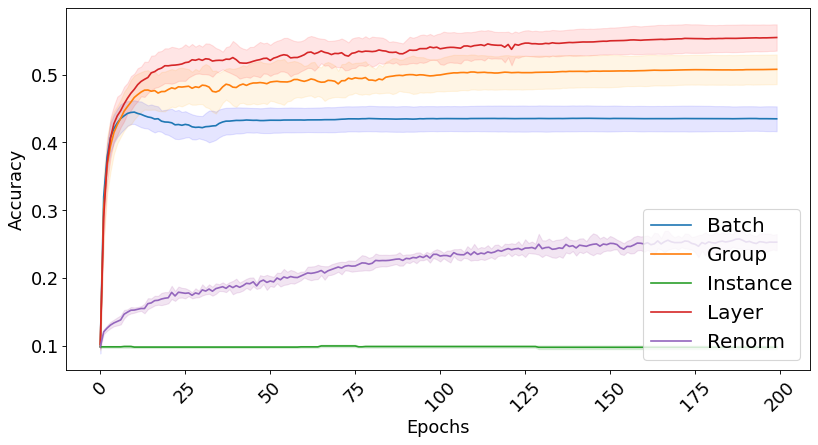}
  }
  \caption{\textbf{Accuracies of ResNet-18 on the uniform and non-IID cases.} BN, GN, and LN require few rounds to reach convergence in uniform setting, while they need more training steps to converge in non-IID scenarios. It is clearly shown that BRN requires a very long learning curve and that IN does not learn in FL.}
  \label{fig:non-iid}
\end{figure*}

This subsection presents the results of the three data partitioning scenarios presented: uniform, labels quantity skew and covariate shift. \Cref{tab:UNIFORM-setting} shows that normalization levels have a huge impact on the performance of a NN, ranging from very poor levels to almost reaching the level of accuracy in the centralized case.
ResNet-18-LN performs slightly better than BN and GN while outperforming IN and BRN in the uniform setting (\cref{subfig-1:uniform}). In both the labels quantity skew and the covariate shift scenarios, both GN and LN outperform all the other normalization layers; however, they require more training steps to converge, as shown in \cref{subfig-2:labels} and \cref{subfig-3:covariate}.
IN does not learn in FL; indeed, since both MNIST and CIFAR10 have ten classes, ResNet-18-IN's performance is like tossing a coin. 
BRN seems to have a very long learning curve; in fact, it needs a lot of training rounds to reach convergence. However, its performance is still far from the best performances of BN, GN and LN. For this reason, the following subsections will report results only for the most promising normalization layers: BN, GN and LN.

\begin{table*}
\centering
\caption{\label{tab:UNIFORM-setting}Accuracy in the uniform setting.\\}
\label{tab:UNIFORM-setting}
\begin{tabular}{lccccc} 
\toprule
\textbf{Dataset}        & \textbf{BN}  & \textbf{GN}  & \textbf{IN} & \textbf{LN} & \textbf{BRN}\\ 

\midrule
MNIST       & $97.12\% \pm 0.24\%$               & $98.26\% \pm 0.19\%$             & $11.28\% \pm 0.04\%$                 & $\textbf{98.51\%} \pm \textbf{0.06\%}$                 & $86.54\% \pm 4.24\%$                 \\
\midrule     
CIFAR10     & $47.12\% \pm 0.82\%$               & $53.99\% \pm 0.23\%$               & $10.04\% \pm 0.07\%$                 & $\textbf{59.06\%} \pm \textbf{0.25\%}$          & $32.36\%  \pm 1.07\%$                \\           
\bottomrule
\end{tabular}
\end{table*}

\begin{table*}
\centering
\caption{\label{tab:LABELSSKEW-setting}Accuracy in the labels quantity skew setting.\\}
\label{tab:LABELSKEW}
\begin{tabular}{lccccc} 
\toprule
\textbf{Dataset}        & \textbf{BN}  & \textbf{GN}  & \textbf{IN} & \textbf{LN} & \textbf{BRN}\\ 

\midrule
MNIST       & $82.25\% \pm 7.70\%$               & $97.32\% \pm 0.46\%$             & $18.53\% \pm 4.18\%$                 & $\textbf{97.68\%} \pm \textbf{0.22\%}$                 & $74.66\% \pm 3.20\%$                 \\
\midrule     
CIFAR10     & $38.02\% \pm 1.64\%$               & $\textbf{58.91\%} \pm \textbf{3.10\%}$               & $19.00\% \pm 5.47\%$                 & $57.98\% \pm 3.88\%$          & $27.34\%  \pm 5.56\%$                \\           
\bottomrule
\end{tabular}
\end{table*}

\begin{table*}
\centering
\caption{\label{tab:COVARIATE-setting}Accuracy in the covariate shift setting.\\}
\label{tab:COVARIATE}
\begin{tabular}{lccccc} 
\toprule
\textbf{Dataset}        & \textbf{BN}  & \textbf{GN}  & \textbf{IN} & \textbf{LN} & \textbf{BRN}\\ 

\midrule
MNIST       & $96.42\% \pm 0.25\%$               & $\textbf{97.96\%} \pm \textbf{0.19\%}$             & $11.51\% \pm 0.23\%$                 & $97.37 \pm 1.63\%$                 & $91.01\% \pm 1.12\%$                 \\
\midrule     
CIFAR10     & $44.58\% \pm 1.96\%$               & $51.04\% \pm 2.83\%$               & $10.08\% \pm 0.23\%$                 & $\textbf{55.54} \pm \textbf{2.22\%}$          & $26.20\%  \pm 1.48\%$                \\           
\bottomrule
\end{tabular}
\end{table*}

\subsection{Normalization Layers and batch size}
\label{ssec:norm-layers-batch-size}

\begin{table}
\centering
\caption{\label{tab:BATCH}Accuracy in the labels quantity skew setting as the batch size varies.\\}
\label{fig:BATCH}
\begin{tabular}{lrccc} 
\toprule
\textbf{Dataset} & \textbf{Batch size}     & \textbf{BN}  & \textbf{GN} & \textbf{LN}\\ 

\midrule
\multirow{7}{*}{MNIST} &  8    & $94.77\%$               & $97.85\%$               & $98.32\%$                    \\
 
& 16        & $81.02\%$               & $97.92\%$               & $98.83\%$                    \\

& 32        & $98.00\%$               & $97.93\%$               & $98.36\%$                    \\ 

& 64        & $73.89\%$               & $95.91\%$               & $98.13\%$                 \\

& 128       & $87.07\%$               & $96.84\%$               & $97.39\%$                \\

& 256       & $90.33\%$               & $97.22\%$               & $97.39\%$                \\

& 512       & $87.86\%$               & $97.97\%$               & $96.99\%$               \\
\midrule
\multirow{7}{*}{CIFAR10} &  8   & $40.23\%$               & $59.95\%$               & $66.63\%$                    \\

& 16        & $45.41\%$               & $61.87\%$               & $56.42\%$                    \\

& 32        & $42.26\%$               & $55.61\%$               & $52.70\%$                    \\ 

& 64        & $43.02\%$               & $59.43\%$               & $55.07\%$                 \\

& 128       & $38.02\%$               & $60.17\%$               & $60.36\%$                \\

& 256       & $46.69\%$               & $56.18\%$               & $56.92\%$                \\

& 512       & $35.86\%$               & $51.92\%$               & $59.73\%$               \\
\bottomrule
\end{tabular}
\end{table}

\begin{figure}[H]
  \includegraphics[width=0.48\textwidth]{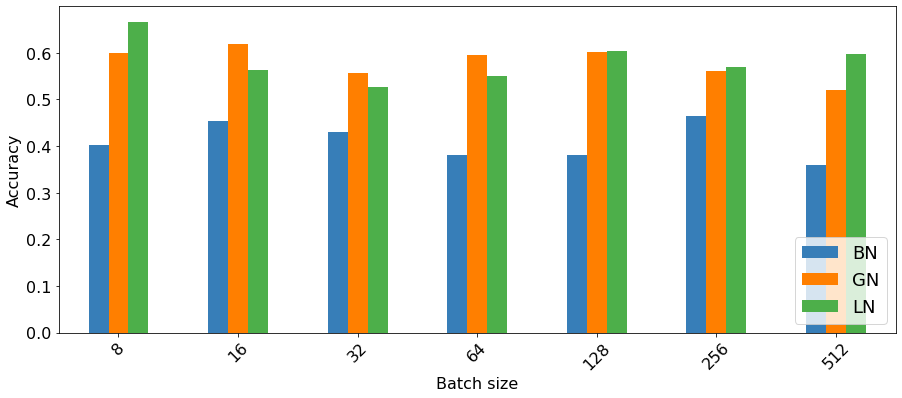}
  \caption{\textbf{Accuracies on CIFAR10 and different batch sizes.} Accuracy degrades when the batch size becomes too large.}
  \label{fig:batch-size}
\end{figure}

We examined the effect of a range of batch sizes on training NNs with different normalization layers (Tab. \ref{tab:BATCH} and Fig.\ref{fig:batch-size}). We trained ResNet-18 on both MNIST and CIFAR10 with a batch size of 8, 16, 32, 64, 128, 256 and 512.

In Tab.\ref{tab:BATCH} we can see that GN and LN variants of ResNet-18 consistently outperform BN (batch sizes 8 and 16). In all three variants, the accuracy degrades when the batch size becomes too large (in almost all cases, there is a significant drop in performance when passing batch size 256 to 512). A possible explanation for this phenomenon is that, as stated in~\cite{keskar2017onlarge}, "the lack of generalization ability is due to the fact that large-batch methods tend to converge to sharp minimizers of the training function". This is especially true in contexts such as FL, where clients have fewer data than in centralized scenarios, and therefore increasing the batch size has a greater effect.

\subsection{Normalization layers and number of epochs per round}
\label{ssec:norm-layers-epochs-round}

We considered two types of experiments to study how accuracy is affected by the number of epochs per round:
\begin{itemize}
	\item Fix the number of rounds and increase the number of local training steps (Tab.\ref{fig:EPOCHS}).
	\item Fix the number of training epochs to 1000 and vary the ratio of epochs to rounds (Tab. \ref{fig:norm-epochs-round}).
\end{itemize}

It can be noted that models take benefit from more local steps of gradient descent before doing aggregation. Indeed, accuracy increases as E increases. A possible explanation is that this happens because clients of the federation share a similar loss function shape, and going more and more towards the local minima can be beneficial to reach global optima. 

Interestingly, when E=1, BN converges quickly, while GN and LN require more training steps to converge. However, when E increases to 10 or 100, BN also requires more rounds to reach convergence, while the learning curves of GN and LN are unaffected by significant changes.

\begin{figure*}[!ht]
  \subfloat[BN\label{subfig-1:epochs_bn}]{%
    \includegraphics[width=0.33\textwidth]{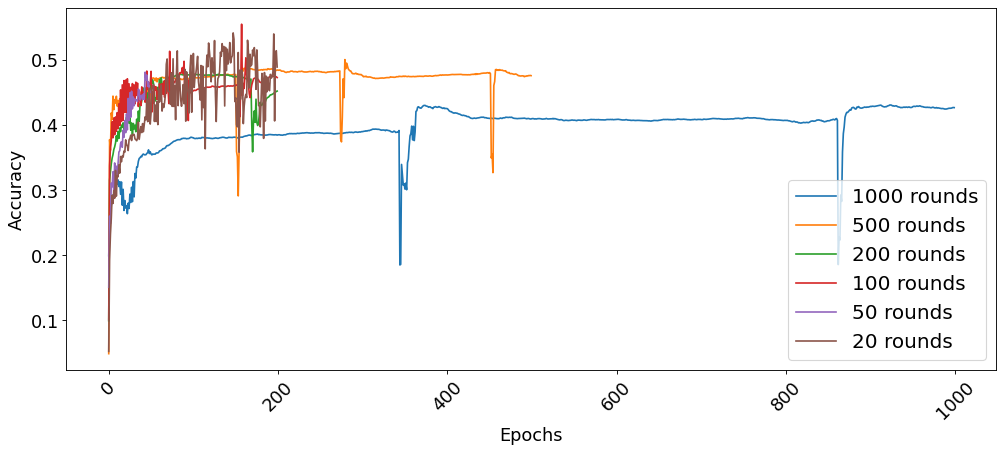}
  }
  \hfill
  \subfloat[GN\label{subfig-2:epochs_gn}]{%
    \includegraphics[width=0.33\textwidth]{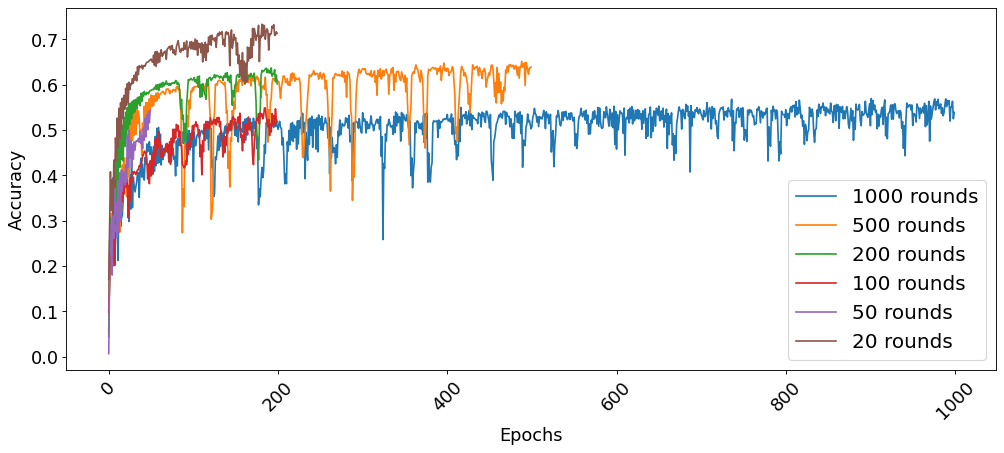}
  }
  \subfloat[LN\label{subfig-3:epochs_ln}]{%
    \includegraphics[width=0.33\textwidth]{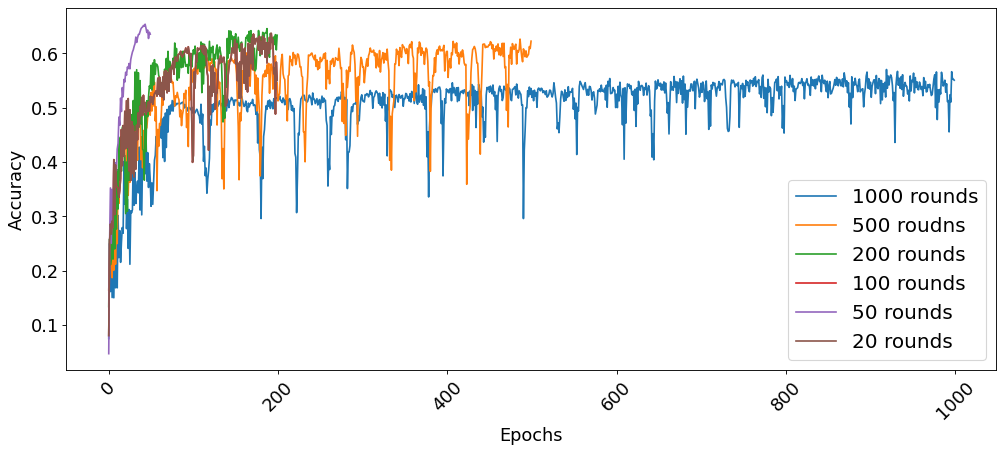}
  }
  \caption{\textbf{Accuracies on CIFAR10 fixing the number of epochs to 1000 and varying the ratio of epochs to rounds.}}
  \label{fig:epochs_fixed}
\end{figure*}

These results can also be analyzed from a communication point of view: with the same amount of epochs, less communication achieves better results. For example, on CIFAR10, ResNet-GN with E=2 and 500 rounds achieves higher accuracy than ResNet-GN with E=1 and 1000 rounds (Fig.\ref{fig:epochs_fixed}). 
This means that perhaps counter-intuitively, training locally before performing aggregation can boost the model’s accuracy. This seems to indicate that pursuing local optimizations can lead to better approximations of the local optima. However, at a certain point, increasing E and reducing the number of rounds decreases the performance. This pattern is clearly visible with all the normalization layers and in both datasets. \Cref{fig:norm-epochs-round} shows that we always need an appropriate ratio of epochs to round.

\begin{table}
\centering
\caption{\label{tab:EPOCHS}Accuracy in the labels quantity skew setting as the number of epochs per round varies.\\}
\label{fig:EPOCHS}
\begin{tabular}{lrcccc} 
\toprule
\textbf{Dataset} & \textbf{Epochs}  & \textbf{BN}  & \textbf{GN} & \textbf{LN}\\ 

\midrule
\multirow{3}{*}{MNIST} &   1       & $87.07\%$               & $96.84\%$               & $97.39\%$                    \\

& 10     & $79.82\%$               & $98.83\%$               & $98.95\%$                  \\

& 100    & $92.88\%$               & $97.83\%$               & $99.04\%$                 \\

\midrule
\multirow{3}{*}{CIFAR10} & 1     & $38.02\%$               & $60.17\%$               & $60.36\%$                    \\

& 10     & $55.53\%$               & $54.89\%$               & $63.68\%$                  \\

& 100    & $54.19\%$               & $73.32\%$               & $68.40\%$                 \\
\bottomrule
\end{tabular}
\end{table}

\begin{table}
\centering
\caption{\label{tab:norm-epoch-round}Comparison between different epochs per rounds in labels quantity skew setting.\\}
\label{fig:norm-epochs-round}
\begin{tabular}{lrcccc}

\toprule
\textbf{Dataset} &  \textbf{Epochs}      &  \textbf{Rounds}                     & \textbf{BN} & \textbf{GN} & \textbf{LN}  \\ 
\midrule
\multirow{7}{*}{MNIST} &
1       & 1000                    & $91.15\%$                & $98.92\%$               & $98.77\%$           \\

& 2       & 500                    & $84.68\%$               & $98.85\%$               & $99.05\%$           \\

& 5        & 200                   & $92.42\%$               & $98.33\%$               & $99.12\%$        \\

& 10        & 100                  & $79.40\%$               & $98.46\%$               & $98.69\%$    \\

& 20        & 50                   & $77.86\%$               & $97.17\%$               & $97.26\%$         \\

& 50        & 20                   & $54.18\%$               & $79.77\%$               & $74.40\%$    \\

& 100        & 10                  & $60.29\%$               & $78.02\%$               & $78.86\%$            \\
\midrule
\multirow{7}{*}{CIFAR10} &
1       & 1000                    & $43.12\%$               & $56.94\%$               & $57.03\%$           \\

& 2       & 500                   & $50.09\%$               & $65.14\%$               & $62.63\%$           \\

& 5        & 200                  & $47.85\%$               & $63.68\%$               & $64.56\%$        \\

& 10        & 100                 & $51.37\%$               & $51.23\%$               & $61.22\%$    \\

& 20        & 50                  & $48.16\%$               & $54.33\%$               & $65.36\%$         \\

& 50        & 20                  & $38.92\%$               & $56.53\%$               & $52.67\%$    \\

& 100        & 10                 & $33.06\%$               & $48.24\%$               & $50.56\%$            \\

\bottomrule
\end{tabular}
\end{table}

\subsection{Normalization layers and number of clients}
\label{ssec:norm-layers-number-clients}

\begin{table}
\centering\caption{\label{tab:clients}Accuracy in the labels quantity skew setting as the number of collaborators varies.\\}
\begin{tabular}{lrccc} 
\toprule
\textbf{Dataset} & \textbf{Clients}  & \textbf{BN}  & \textbf{GN} & \textbf{LN}\\ 

\midrule
\multirow{4}{*}{MNIST} &   2         & $99.02\%$               & $99.61\%$               & $99.85\%$                    \\

& 4    & $91.85\%$               & $99.47\%$               & $98.36\%$                 \\

& 8    & $88.38\%$               & $96.49\%$               & $97.68\%$                \\

& 10   & $87.07\%$               & $96.84\%$               & $97.39\%$            \\
                           
\midrule
\multirow{4}{*}{CIFAR10} & 2         & $44.57\%$               & $74.15\%$               & $75.63\%$                    \\

& 4    & $55.30\%$               & $50.50\%$               & $67.69\%$                 \\

& 8    & $51.27\%$               & $60.75\%$               & $65.00\%$                \\

& 10   & $38.02\%$               & $60.17\%$               & $60.36\%$                \\
\bottomrule
\end{tabular}
\end{table}

\begin{figure}[H]
  \includegraphics[width=0.48\textwidth]{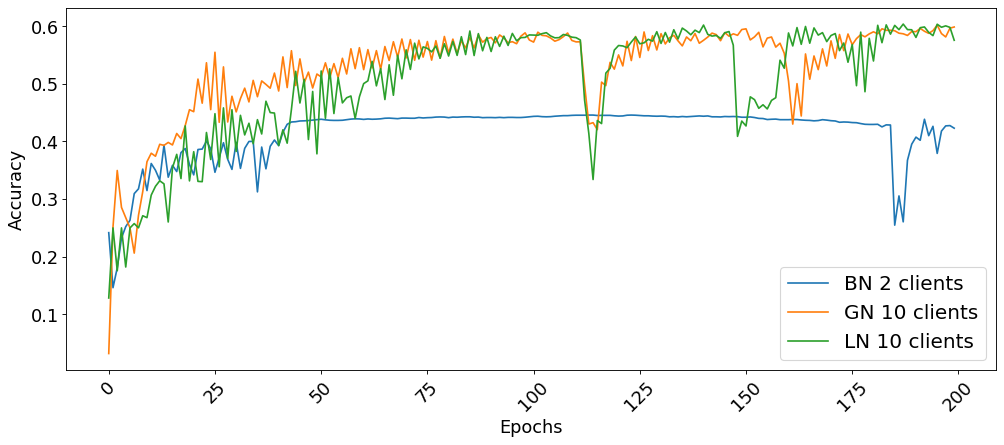}
  \caption{\textbf{Accuracies on CIFAR10 and different number of clients.} Accuracy of ResNet-18-BN on 2 parties is still lower than ResNet-18-GN/LN on 10 parties.}
  \label{fig:CLIENTS}
\end{figure}

We tested the scalability of FL by measuring the effect of the number of clients of the federation, as shown in Fig.~\ref{fig:CLIENTS}, and considering two types of experiments:
\begin{itemize}
	\item a labels quantity skew split of the dataset across a different number of clients (namely 2, 4, 8 and 10). Results are reported in \Cref{tab:clients}. 
	\item a uniform dataset split across clients, but considering only some parties. Here the idea is to show how increasing the number of participants, and so the quantity of data, can be beneficial to the federation. Results are reported in \Cref{tab:uniform_clients}.
\end{itemize}

We can observe (\Cref{tab:clients}) that the accuracy significantly increases when decreasing the number of clients. Indeed, when the number of parties is small, the amount of local data increases, leading to better local models, and also aggregating fewer models can result in less information loss. Moreover, we can note the importance of normalization layers in FL: GN and LN variants of ResNet-18 in the ten clients scenario perform better than BN in a two clients scenario on CIFAR10, while on MNIST there is only a slightly drop. \\
\Cref{tab:uniform_clients} shows the results in an IID scenario considering only some shards of the dataset. In this case, the amount of local data remains the same in each configuration; however, the federation's total amount of data varies according to the number of parties. Increasing the quantity of data in the federation by increasing the number of clients benefits the aggregated model. 

\begin{table}
\centering\caption{\label{tab:uniform_clients}Accuracy in the \textit{i.i.d.} setting using only some shards of the dataset.\\}
\begin{tabular}{lrccc} 
\toprule
\textbf{Dataset} & \textbf{Clients}  & \textbf{BN}  & \textbf{GN} & \textbf{LN}\\ 

\midrule
\multirow{4}{*}{MNIST} &   2         & $94.64\%$               & $96.26\%$               & $97.38\%$                    \\

& 4    & $96.18\%$               & $97.52\%$               & $98.10\%$                 \\

& 8    & $96.79\%$               & $97.98\%$               & $98.34\%$                \\

& 10   & $97.12\%$               & $98.26\%$               & $98.51\%$            \\
                           
\midrule
\multirow{4}{*}{CIFAR10} & 2         & $41.80\%$               & $45.37\%$               & $51.73\%$                    \\

& 4    & $43.47\%$               & $48.74\%$               & $54.55\%$                 \\

& 8    & $45.81\%$               & $53.18\%$               & $58.36\%$                \\

& 10   & $47.12\%$               & $53.99\%$           & $59.06\%$                \\
\bottomrule
\end{tabular}
\end{table}

\section{Conclusions}
\label{sec:conclusions}
This work aims to improve the effectiveness of federated learning, focusing on hyper-parameter optimization, starting from understanding which hyper-parameters specifically affect the training of a federated model differently from centralized training. We specifically focused on layer normalization, which is also a hyper-parameter of centralized training, and frequency of model aggregation, which is not an issue in centralized training. 

We experimented with two network architectures and five normalization layers on two public image datasets. We tested Batch, Group, Instance, Layer Normalization and Batch Renormalization in the uniform, label quantity skew and covariate shift settings. Although BN is the state-of-the-art for classical ML techniques, in our experiments, GN and LN outperformed the other normalization layers in all the FL partitioning strategies. 

Through extensive experimentation, we analyzed how the batch size, the number of epochs per round, the number of rounds and the number of clients of the federation affect the aggregated model performance. These additional tests have been conducted in the labels quantity skew scenario, which is the most challenging for FL algorithms, considering the best three normalization layers: BN, GN and LN. 

GN and LN outperform BN in almost all the tests. Results show that regardless of the batch size, GN and LN consistently outperform BN, although batch size affects the model's performance in all cases. Unexpectedly, we observed that the plot of the quality of the model against the frequency of model aggregation (epochs per round) consistently exhibits a maximum at a few epochs per round. For FL, the number of epochs per round exhibits similar behaviour of batch size for centralized training. 

Eventually, we tested the scalability of FL systems. We noted that FL is not scalable under strong scalability assumption, i.e. increasing the number of clients while maintaining constant the size of local datasets. However, GN and LN on ten clients still outperform BN on two clients. The scalability has also been tested in the IID scenario under the weak scalability assumption, i.e. increasing the number of clients while maintaining constant the size of the local dataset per client. In this case, the federation's data changes with the number of clients, and the model's performance increases with the number of parties.

\section*{Acknowledgements}
\label{sec:acknowledgements}
This work has been supported by the Spoke "FutureHPC $\&$ BigData” of the ICSC – Centro Nazionale di Ricerca in "High Performance Computing, Big Data and Quantum Computing", funded by European Union – NextGenerationEU, and by the European Union within the H2020 RIA “European Processor Initiative—Specific Grant Agreement 2” G.A. 826647, \href{https://www.european-processor-initiative.eu/}{https://www.european-processor-initiative.eu/}.

\bibliographystyle{IEEEtran}
\bibliography{main}

\begin{thebibliography}{10}
\providecommand{\url}[1]{#1}
\csname url@samestyle\endcsname
\providecommand{\newblock}{\relax}
\providecommand{\bibinfo}[2]{#2}
\providecommand{\BIBentrySTDinterwordspacing}{\spaceskip=0pt\relax}
\providecommand{\BIBentryALTinterwordstretchfactor}{4}
\providecommand{\BIBentryALTinterwordspacing}{\spaceskip=\fontdimen2\font plus
\BIBentryALTinterwordstretchfactor\fontdimen3\font minus
  \fontdimen4\font\relax}
\providecommand{\BIBforeignlanguage}[2]{{%
\expandafter\ifx\csname l@#1\endcsname\relax
\typeout{** WARNING: IEEEtran.bst: No hyphenation pattern has been}%
\typeout{** loaded for the language `#1'. Using the pattern for}%
\typeout{** the default language instead.}%
\else
\language=\csname l@#1\endcsname
\fi
#2}}
\providecommand{\BIBdecl}{\relax}
\BIBdecl

\bibitem{kairouz2021advances}
P.~Kairouz, H.~B. McMahan, B.~Avent, A.~Bellet, M.~Bennis, A.~N. Bhagoji, K.~A.
  Bonawitz, Z.~Charles, G.~Cormode, R.~Cummings, R.~G.~L. D'Oliveira,
  H.~Eichner, S.~E. Rouayheb, D.~Evans, J.~Gardner, Z.~Garrett,
  A.~Gasc{\'{o}}n, B.~Ghazi, P.~B. Gibbons, M.~Gruteser, Z.~Harchaoui, C.~He,
  L.~He, Z.~Huo, B.~Hutchinson, J.~Hsu, M.~Jaggi, T.~Javidi, G.~Joshi,
  M.~Khodak, J.~Kone{\v{c}}n{\'y}, A.~Korolova, F.~Koushanfar, S.~Koyejo,
  T.~Lepoint, Y.~Liu, P.~Mittal, M.~Mohri, R.~Nock, A.~{\"{O}}zg{\"{u}}r,
  R.~Pagh, H.~Qi, D.~Ramage, R.~Raskar, M.~Raykova, D.~Song, W.~Song, S.~U.
  Stich, Z.~Sun, A.~T. Suresh, F.~Tram{\`{e}}r, P.~Vepakomma, J.~Wang,
  L.~Xiong, Z.~Xu, Q.~Yang, F.~X. Yu, H.~Yu, and S.~Zhao, ``Advances and open
  problems in federated learning,'' \emph{Found. Trends Mach. Learn.}, vol.~14,
  no. 1-2, pp. 1--210, 2021.

\bibitem{callaghan2020data}
S.~Callaghan, ``Data sharing in a time of pandemic,'' \emph{Patterns}, vol.~1,
  no.~5, p. 100086, 2020.

\bibitem{voigt2017eu}
P.~Voigt and A.~v.~d. Bussche, \emph{The EU General Data Protection Regulation
  (GDPR): A Practical Guide}, 1st~ed.\hskip 1em plus 0.5em minus 0.4em\relax
  Springer Publishing Company, Incorporated, 2017.

\bibitem{mcmahan2017communication}
B.~McMahan, E.~Moore, D.~Ramage, S.~Hampson, and B.~A. y~Arcas,
  ``Communication-efficient learning of deep networks from decentralized
  data,'' in \emph{Proc. of the 20th Intl. Conference on Artificial
  Intelligence and Statistics, {AISTATS} 2017, 20-22 April 2017, Fort
  Lauderdale, FL, {USA}}, ser. Proc. of Machine Learning Research, A.~Singh and
  X.~J. Zhu, Eds., vol.~54.\hskip 1em plus 0.5em minus 0.4em\relax {PMLR},
  2017, pp. 1273--1282.

\bibitem{polato2022boosting}
M.~Polato, R.~Esposito, and M.~Aldinucci, ``Boosting the federation: Cross-silo
  federated learning without gradient descent,'' in \emph{International Joint
  Conference on Neural Networks, {IJCNN} 2022, Padua, Italy, July 18-23,
  2022}.\hskip 1em plus 0.5em minus 0.4em\relax {IEEE}, 2022, pp. 1--10.

\bibitem{23:praise-fl:pdp}
Y.~Arfat, G.~Mittone, I.~Colonnelli, F.~D'Ascenzo, R.~Esposito, and
  M.~Aldinucci, ``Pooling critical datasets with federated learning,'' in
  \emph{Proc. of 31st Euromicro Intl. Conference on Parallel Distributed and
  network-based Processing (PDP)}.\hskip 1em plus 0.5em minus 0.4em\relax
  Napoli, Italy: IEEE, 2023.

\bibitem{li2019convergence}
X.~Li, K.~Huang, W.~Yang, S.~Wang, and Z.~Zhang, ``On the convergence of fedavg
  on non-iid data,'' \emph{CoRR}, vol. abs/1907.02189, 2019.

\bibitem{li2020federated}
T.~Li, A.~K. Sahu, M.~Zaheer, M.~Sanjabi, A.~Talwalkar, and V.~Smith,
  ``Federated optimization in heterogeneous networks,'' in \emph{Proc. of
  Machine Learning and Systems 2020, MLSys 2020, Austin, TX, USA, March 2-4,
  2020}, I.~S. Dhillon, D.~S. Papailiopoulos, and V.~Sze, Eds.\hskip 1em plus
  0.5em minus 0.4em\relax mlsys.org, 2020.

\bibitem{wang2020tackling}
J.~Wang, Q.~Liu, H.~Liang, G.~Joshi, and H.~V. Poor, ``Tackling the objective
  inconsistency problem in heterogeneous federated optimization,'' in
  \emph{Advances in Neural Information Processing Systems 33: Annual Conference
  on Neural Information Processing Systems 2020, NeurIPS 2020, December 6-12,
  2020, virtual}, H.~Larochelle, M.~Ranzato, R.~Hadsell, M.~Balcan, and H.~Lin,
  Eds., 2020.

\bibitem{karimireddy2020scaffold}
S.~P. Karimireddy, S.~Kale, M.~Mohri, S.~J. Reddi, S.~U. Stich, and A.~T.
  Suresh, ``{SCAFFOLD:} stochastic controlled averaging for federated
  learning,'' in \emph{Proc. of the 37th Intl. Conference on Machine Learning,
  {ICML} 2020, 13-18 July 2020, Virtual Event}, ser. Proc. of Machine Learning
  Research, vol. 119.\hskip 1em plus 0.5em minus 0.4em\relax {PMLR}, 2020, pp.
  5132--5143.

\bibitem{shoham2019overcoming}
N.~Shoham, T.~Avidor, A.~Keren, N.~Israel, D.~Benditkis, L.~Mor{-}Yosef, and
  I.~Zeitak, ``Overcoming forgetting in federated learning on non-iid data,''
  \emph{CoRR}, vol. abs/1910.07796, 2019.

\bibitem{li2022federated}
Q.~Li, Y.~Diao, Q.~Chen, and B.~He, ``Federated learning on non-iid data silos:
  An experimental study,'' in \emph{38th {IEEE} International Conference on
  Data Engineering, {ICDE} 2022, Kuala Lumpur, Malaysia, May 9-12, 2022}.\hskip
  1em plus 0.5em minus 0.4em\relax {IEEE}, 2022, pp. 965--978.

\bibitem{casella2022benchmarking}
B.~Casella, R.~Esposito, C.~Cavazzoni, and M.~Aldinucci, ``Benchmarking fedavg
  and fedcurv for image classification tasks,'' in \emph{Proc. of th 1st
  Italian Conference on Big Data and Data Science, {ITADATA22}, Milan,
  September 20-21, 2022}, 2022.

\bibitem{ioffe2015batch}
S.~Ioffe and C.~Szegedy, ``Batch normalization: Accelerating deep network
  training by reducing internal covariate shift,'' in \emph{Proceedings of the
  32nd International Conference on Machine Learning, {ICML} 2015, Lille,
  France, 6-11 July 2015}, ser. {JMLR} Workshop and Conference Proceedings,
  F.~R. Bach and D.~M. Blei, Eds., vol.~37.\hskip 1em plus 0.5em minus
  0.4em\relax JMLR.org, 2015, pp. 448--456.

\bibitem{lecun1998gradient}
Y.~LeCun, L.~Bottou, Y.~Bengio, and P.~Haffner, ``Gradient-based learning
  applied to document recognition,'' \emph{Proc. {IEEE}}, vol.~86, no.~11, pp.
  2278--2324, 1998.

\bibitem{krizhevsky2009learning}
A.~Krizhevsky, G.~Hinton \emph{et~al.}, ``Learning multiple layers of features
  from tiny images,'' pp. 32--33, 2009.

\bibitem{yang2021medmnist}
J.~Yang, R.~Shi, D.~Wei, Z.~Liu, L.~Zhao, B.~Ke, H.~Pfister, and B.~Ni,
  ``Medmnist v2: {A} large-scale lightweight benchmark for 2d and 3d biomedical
  image classification,'' \emph{CoRR}, vol. abs/2110.14795, 2021.

\bibitem{wu2018group}
Y.~Wu and K.~He, ``Group normalization,'' in \emph{Computer Vision - {ECCV}
  2018 - 15th European Conference, Munich, Germany, September 8-14, 2018,
  Proceedings, Part {XIII}}, ser. Lecture Notes in Computer Science,
  V.~Ferrari, M.~Hebert, C.~Sminchisescu, and Y.~Weiss, Eds., vol. 11217.\hskip
  1em plus 0.5em minus 0.4em\relax Springer, 2018, pp. 3--19.

\bibitem{ba2016layer}
L.~J. Ba, J.~R. Kiros, and G.~E. Hinton, ``Layer normalization,'' \emph{CoRR},
  vol. abs/1607.06450, 2016.

\bibitem{ulyanov2016instance}
D.~Ulyanov, A.~Vedaldi, and V.~S. Lempitsky, ``Instance normalization: The
  missing ingredient for fast stylization,'' \emph{CoRR}, vol. abs/1607.08022,
  2016.

\bibitem{ioffe2017batch}
S.~Ioffe, ``Batch renormalization: Towards reducing minibatch dependence in
  batch-normalized models,'' in \emph{Advances in Neural Information Processing
  Systems 30: Annual Conference on Neural Information Processing Systems 2017,
  December 4-9, 2017, Long Beach, CA, {USA}}, I.~Guyon, U.~von Luxburg,
  S.~Bengio, H.~M. Wallach, R.~Fergus, S.~V.~N. Vishwanathan, and R.~Garnett,
  Eds., 2017, pp. 1945--1953.

\bibitem{casella2022transfer}
B.~Casella, A.~B. Chisari, S.~Battiato, and M.~V. Giuffrida, ``Transfer
  learning via test-time neural networks aggregation,'' in \emph{Proceedings of
  the 17th International Joint Conference on Computer Vision, Imaging and
  Computer Graphics Theory and Applications, {VISIGRAPP} 2022, Volume 5:
  VISAPP, Online Streaming, February 6-8, 2022}, G.~M. Farinella, P.~Radeva,
  and K.~Bouatouch, Eds.\hskip 1em plus 0.5em minus 0.4em\relax {SCITEPRESS},
  2022, pp. 642--649.

\bibitem{andreux2020siloed}
M.~Andreux, J.~O. du~Terrail, C.~Beguier, and E.~W. Tramel, ``Siloed federated
  learning for multi-centric histopathology datasets,'' in \emph{Domain
  Adaptation and Representation Transfer, and Distributed and Collaborative
  Learning - Second {MICCAI} Workshop, {DART} 2020, and First {MICCAI}
  Workshop, {DCL} 2020, Held in Conjunction with {MICCAI} 2020, Lima, Peru,
  October 4-8, 2020, Proceedings}, ser. Lecture Notes in Computer Science,
  S.~Albarqouni, S.~Bakas, K.~Kamnitsas, M.~J. Cardoso, B.~A. Landman, W.~Li,
  F.~Milletari, N.~Rieke, H.~Roth, D.~Xu, and Z.~Xu, Eds., vol. 12444.\hskip
  1em plus 0.5em minus 0.4em\relax Springer, 2020, pp. 129--139.

\bibitem{li2021fedbn}
X.~Li, M.~Jiang, X.~Zhang, M.~Kamp, and Q.~Dou, ``Fedbn: Federated learning on
  non-iid features via local batch normalization,'' in \emph{9th International
  Conference on Learning Representations, {ICLR} 2021, Virtual Event, Austria,
  May 3-7, 2021}.\hskip 1em plus 0.5em minus 0.4em\relax OpenReview.net, 2021.

\bibitem{he2016deep}
K.~He, X.~Zhang, S.~Ren, and J.~Sun, ``Deep residual learning for image
  recognition,'' in \emph{2016 {IEEE} Conference on Computer Vision and Pattern
  Recognition, {CVPR} 2016, Las Vegas, NV, USA, June 27-30, 2016}.\hskip 1em
  plus 0.5em minus 0.4em\relax {IEEE} Computer Society, 2016, pp. 770--778.

\bibitem{reina2021openfl}
G.~A. Reina, A.~Gruzdev, P.~Foley, O.~Perepelkina, M.~Sharma, I.~Davidyuk,
  I.~Trushkin, M.~Radionov, A.~Mokrov, D.~Agapov, J.~Martin, B.~Edwards, M.~J.
  Sheller, S.~Pati, P.~N. Moorthy, H.~S. Wang, P.~Shah, and S.~Bakas, ``Openfl:
  An open-source framework for federated learning,'' \emph{CoRR}, vol.
  abs/2105.06413, 2021.

\bibitem{tan2019efficientnet}
M.~Tan and Q.~V. Le, ``Efficientnet: Rethinking model scaling for convolutional
  neural networks,'' in \emph{Proceedings of the 36th International Conference
  on Machine Learning, {ICML} 2019, 9-15 June 2019, Long Beach, California,
  {USA}}, ser. Proceedings of Machine Learning Research, K.~Chaudhuri and
  R.~Salakhutdinov, Eds., vol.~97.\hskip 1em plus 0.5em minus 0.4em\relax
  {PMLR}, 2019, pp. 6105--6114.

\bibitem{moreau2022benchopt}
T.~Moreau, M.~Massias, A.~Gramfort, P.~Ablin, P.~Bannier, B.~Charlier,
  M.~Dagr{\'{e}}ou, T.~D. la~Tour, G.~Durif, C.~F. Dantas, Q.~Klopfenstein,
  J.~Larsson, E.~Lai, T.~Lefort, B.~Mal{\'{e}}zieux, B.~Moufad, B.~T. Nguyen,
  A.~Rakotomamonjy, Z.~Ramzi, J.~Salmon, and S.~Vaiter, ``Benchopt:
  Reproducible, efficient and collaborative optimization benchmarks,''
  \emph{CoRR}, vol. abs/2206.13424, 2022.

\bibitem{keskar2017onlarge}
N.~S. Keskar, D.~Mudigere, J.~Nocedal, M.~Smelyanskiy, and P.~T.~P. Tang, ``On
  large-batch training for deep learning: Generalization gap and sharp
  minima,'' in \emph{5th International Conference on Learning Representations,
  {ICLR} 2017, Toulon, France, April 24-26, 2017, Conference Track
  Proceedings}.\hskip 1em plus 0.5em minus 0.4em\relax OpenReview.net, 2017.

\end{thebibliography}

\begin{IEEEbiographynophoto}{Bruno Casella}
is currently a PhD student in Modeling and Data Science, at the University of Turin.
He received the B.S. degree in Computer Engineering in 2020, and the M. Sc. in Data Science for management from the University of Catania. His current research interests include federated learning, transfer learning and distributed computing.
\end{IEEEbiographynophoto}

x\begin{IEEEbiographynophoto}{Roberto Esposito}
is an associate professor at the University of Turin. He received the PhD in Computer Science in 2003 from the University of Turin.

\end{IEEEbiographynophoto}

\begin{IEEEbiographynophoto}{Antonio Sciarappa}
is currently a researcher at the Leonardo Labs of Genova. He received the PhD in Theoretical Particle Physics in 2015 from the SISSA. His current research interests include parallel and distributed computing.

\end{IEEEbiographynophoto}

\begin{IEEEbiographynophoto}{Carlo Cavazzoni}
is currently Senior Vice President of Cloud Computing and Head of the High Performance Computing of Leonardo S.p.A. He received the PhD in Computational Physics from the SISSA.

\end{IEEEbiographynophoto}

\begin{IEEEbiographynophoto}{Marco Aldinucci}
is a full professor and P.I. of the Parallel Computing research group at the University of Turin. He received his PhD from the University of Pisa (2003), and he has been a researcher at the Italian National Research Council (CNR).

\end{IEEEbiographynophoto}

\section*{Appendix}
\label{sec:appendix}
\noindent Accuracies on the uniform and non-iid data settings using and EfficientNet-B0~\cite{tan2019efficientnet}.

\begin{table}[H]
\caption{\label{tab:UNIFORM-efficient}Accuracy in the uniform setting with EfficientNet-B0\\}
\label{tab:UNIFORM-efficient}
\begin{tabular}{lccccc} 
\toprule
\textbf{Dataset}   & \textbf{BN}  & \textbf{GN}  & \textbf{IN} & \textbf{LN} & \textbf{BRN}\\ 

\midrule
MNIST     & $65.09\%$               & $95.07\%$             & $11.26\%$                 & $96.12\%$          & $11.02\%$                 \\
\midrule     
CIFAR10   & $29.88\%$               & $29.93\%$            & $10.08\%$                 & $38.98\%$          & $10.48\%$                \\     
\bottomrule
\end{tabular}
\end{table}

\begin{table}[H]
\caption{\label{tab:LABELSSKEW-efficient}Accuracy in the labels quantity skew setting with EfficientNet-B0\\}
\label{tab:LABELSKEW-efficient}
\begin{tabular}{lccccc} 
\toprule
\textbf{Dataset}   & \textbf{BN}  & \textbf{GN}  & \textbf{IN} & \textbf{LN} & \textbf{BRN}\\ 

\midrule
MNIST     & $55.83\%$               & $91.91\%$             & $24.13\%$                 & $94.35\%$          & $45.69\%$                 \\
\midrule     
CIFAR10   & $29.98\%$               & $27.53\%$            & $14.92\%$                 & $37.71\%$          & $22.61\%$                \\     
\bottomrule
\end{tabular}
\end{table}

\begin{table}[H]
\caption{\label{tab:COVARIATE-efficient}Accuracy in the covariate shift setting with EfficientNet-B0\\}
\label{tab:COVARIATE-efficient}
\begin{tabular}{lccccc} 
\toprule
\textbf{Dataset}   & \textbf{BN}  & \textbf{GN}  & \textbf{IN} & \textbf{LN} & \textbf{BRN}\\ 

\midrule
MNIST     & $65.10\%$               & $94.86\%$             & $11.19\%$                 & $95.81\%$          & $61.90\%$                 \\
\midrule     
CIFAR10   & $26.78\%$               & $28.01\%$            & $10.16\%$                 & $48.10\%$          & $10.01\%$                \\     
\bottomrule
\end{tabular}
\end{table}

\end{document}